\documentclass[11pt]{article}

\usepackage[final]{acl}

\usepackage{times}
\usepackage{latexsym}

\usepackage[T1]{fontenc}

\usepackage[utf8]{inputenc}

\usepackage{microtype}

\usepackage{inconsolata}

\usepackage{graphicx}
\usepackage{booktabs}       
\usepackage[most]{tcolorbox}
\usepackage{dashrule}
\usepackage{caption}
\usepackage{makecell}
\usepackage{listings}

\usepackage{tabularx}

\title{Ask or Assume? Uncertainty-Aware Clarification-Seeking in Coding Agents}

\author{
  Nicholas Edwards\textsuperscript{1,2} \qquad Sebastian Schuster\textsuperscript{1} \\
  \\
  \textsuperscript{1}Faculty of Computer Science, University of Vienna, Vienna, Austria\\
  \textsuperscript{2}UniVie Doctoral School Computer Science, University of Vienna, Vienna, Austria\\
  \texttt{\{nicholas.edwards, sebastian.schuster\}@univie.ac.at}
}

\begin{document}
\maketitle
\begin{abstract}
As Large Language Model (LLM) agents are increasingly deployed in open-ended domains like software engineering, they frequently encounter \textit{underspecified} instructions that lack crucial context. While human developers naturally resolve underspecification by asking clarifying questions, current agents are largely optimized for autonomous execution. In this work, we systematically evaluate the clarification-seeking abilities of LLM agents on an underspecified variant of SWE-bench Verified. We propose an uncertainty-aware multi-agent scaffold that decouples underspecification detection from code execution. Across both proprietary and open-weight frontier LLMs, our scaffold achieves a 69.40\% task resolve rate, significantly outperforming a standard single-agent setup and closing the performance gap with agents operating on fully specified instructions. Furthermore, we find that the multi-agent system exhibits well-calibrated information-seeking behavior, conserving queries on simple tasks while proactively seeking information on more complex issues. These findings indicate that current models can be turned into proactive collaborators, where agents independently recognize when to ask questions to elicit missing information in real-world, underspecified tasks.
\end{abstract}

\section{Introduction}
A core property of a good collaborator is the ability to establish and maintain a shared understanding of a goal \citep{clark1986referring}. In human communication, this is rarely achieved through a single perfectly specified instruction; instead, interlocutors infer missing context and explicitly signal uncertainty, such as by asking clarification questions \citep{clark1996using, hawkins2015you}.

While information gathering in response to underspecification has been studied in the NLP domain, such as in question answering and conversational search \citep{rao-daume-iii-2018-learning, aliannejadi2019asking, zhang-choi-2025-clarify}, these settings involve static or single-turn interactions over bounded contexts.

In contrast, modern AI agents are increasingly deployed in more naturalistic, open-ended domains such as software engineering. For instance, fixing real-world GitHub issues \citep{jimenez2024swebench} requires exploring, understanding, and editing large-scale repositories often containing hundreds of files and thousands of lines of code, while remaining aligned with implicit developer intentions. Existing agents are optimized for autonomous completion rather than interactive collaboration, creating a critical gap between user intent and agent execution \citep{research-update-algorithmic-vs-holistic-evaluation, shen-etal-2026-scaling, wangposition}.

Given these constraints, we investigate the abilities of LLMs to assess uncertainty and determine how and when to seek information. Specifically, we consider how models handle \textbf{\textit{external uncertainty}}: uncertainty arising from ambiguous or underspecified task inputs, as opposed to internal uncertainty inherent to the model itself \citep{Li_Yu_Zhang_Zhuang_Shah_Sadagopan_Beniwal_2026}. While LLMs can exhibit sensitivity to uncertainty \citep{kadavath2022language}, this is not always robust \citep{andukuri2024stargate, kapoor2024large}. Beyond simply detecting external uncertainty, a true agentic collaborator must be able to continuously monitor for such uncertainty and proactively initiate dialogue to elicit missing information. We define such a system as \textit{uncertainty-aware}. Importantly, we use this term in a behavioral rather than formalized mathematical sense, evaluating how effectively an agent modulates its information-seeking behavior based on the severity of underspecification.

We systematically evaluate the clarification-seeking abilities of LLM agents within a dynamic, multi-turn software engineering framework. We conduct our evaluation using an underspecified variant of SWE-bench Verified \citep{chowdhury2024swebenchverified, vijayvargiya2026interactive}, where information is removed from the dataset's original GitHub issues. To assess the capabilities of agents to act as question-asking collaborators, we design agent scaffolds where models must rely on their own calibration of external uncertainty to decide when to query a user.

\paragraph{Contributions} We develop and test both single- and multi-agent frameworks using frontier LLMs (Claude Sonnet 4.5 and Kimi K2.6) to assess how agents handle uncertainty arising from underspecified inputs. We find that interactive agents can successfully identify and retrieve missing information, obtaining a resolve rate comparable to an autonomous agent provided with a fully specified issue. Our results provide evidence that some models exhibit well-calibrated clarification-seeking behavior, accurately recognizing when an issue is already resolvable and refraining from unnecessary interaction. Our code is available at \href{https://github.com/nedwards99/ask-or-assume}{https://github.com/nedwards99/ask-or-assume}.

\section{Related Work}

\paragraph{Proactive Dialogue Agents} In human communication, interlocutors use clarification requests to establish shared understanding and resolve ambiguity \citep{clark1986referring}. Recent work in the NLP domain has considered clarification-seeking behavior in the context of dialogue-based tasks like open-domain question-answering, conversational search and preference elicitation \citep{rao-daume-iii-2018-learning, aliannejadi2019asking, kim-etal-2024-aligning, li2025eliciting, zhang-choi-2025-clarify, zhang2025modeling}.

To elicit these proactive clarification-seeking behaviors, prior work has explored various methodological approaches. For instance, \citet{zhang2025modeling} propose modeling the utility of future conversation trajectories to teach LLMs when a question is necessary; \citet{andukuri2024stargate} utilize simulated self-play to iteratively improve the model's clarification skills; and \citet{testoni-fernandez-2024-asking} introduce an uncertainty-guided approach, demonstrating that leveraging a model's own uncertainty (operationalized as predictive entropy) to trigger questions is more effective than imitating human clarification-seeking behavior. While these methods have proven effective in bounded conversational settings, our work extends this paradigm by evaluating how agents dynamically elicit missing information in open-ended, multi-step software engineering environments.

\paragraph{Agentic Systems for Ambiguous Code Generation} As LLM agents are increasingly deployed in software engineering, robustly handling ambiguous or underspecified instructions presents a major challenge. Existing coding environments used to evaluate clarification-seeking behaviors often focus on tasks targeting edits to isolated functions, typically allowing only a single, predetermined round of clarification or short, constrained exchanges \citep{li-etal-2023-python, mu2024clarifygpt, wu2025humanevalcomm, li2026clarevalbenchmarkevaluatingclarification}.

To improve the clarification-seeking behavior of agents in dynamic, multi-turn coding settings, recent literature has introduced training pipelines leveraging reinforcement learning \citep{sun2026training} and contrastive training \citep{chen2025learning} techniques. However, these frameworks often impose rigid interaction structures. For example, both \citet{vijayvargiya2026asking} and \citet{darji2025curiositydesignllmbasedcoding} strictly constrain question-asking to the very beginning of the trajectory, deferring all code execution until the initial prompt is disambiguated.

In practice, relevant missing information may only become apparent after initial exploration of a codebase. Most closely related to our evaluation setting, \citet{vijayvargiya2026interactive} analyze how LLM agents handle underspecification in software engineering tasks. In our work, we demonstrate that a specialized multi-agent scaffold  that decouples underspecification detection from code execution enables agents to navigate underspecified tasks and actively ask questions throughout a trajectory.

\paragraph{Uncertainty Detection in LLMs} A prerequisite for proactive clarification is an agent's ability to accurately gauge its own uncertainty. Foundational work demonstrates that base LLMs exhibit a degree of internal calibration, enabling them to estimate the likelihood of their correctness or express their confidence in natural language \citep{kadavath2022language, lin2022teaching}. However, this calibration is not always robust out of the box \citep{kapoor2024large, andukuri2024stargate}, as standard post-training alignment techniques, such as RLHF, can inadvertently degrade calibration and encourage model overconfidence \citep{zhou-etal-2024-relying}. Furthermore, effective interaction requires distinguishing between different sources of uncertainty. For instance, recent work has highlighted the difference between uncertainty caused by a model's lack of parametric knowledge versus input underspecification or ambiguity \citep{10.5555/3692070.3692835, zhang-choi-2025-clarify}. Formally quantifying this distinction, \citet{Li_Yu_Zhang_Zhuang_Shah_Sadagopan_Beniwal_2026} distinguish \textit{internal} uncertainty (inherent to the model) from \textit{external} uncertainty (arising from ambiguous or underspecified inputs). Our work specifically investigates how agents handle external uncertainty. Rather than considering a formalized measure of confidence (e.g., a probability), our work evaluates uncertainty detection behaviorally. That is, we assess whether agents can reliably identify missing context and resolve it through asking the user for clarification.

\section{Method}
\subsection{Dataset and Evaluation Framework}
\label{sec:dataset}
We closely follow and adapt the interactive evaluation setting introduced by \citet{vijayvargiya2026interactive}. We use their dataset \citep{interactive-swe-bench}, which is an underspecified variant of SWE-bench Verified \citep{chowdhury2024swebenchverified}, a human-annotated subset of 500 GitHub issues derived from the original SWE-bench dataset \citep{jimenez2024swebench}. To construct their dataset, the authors utilized GPT-4o \citep{openai2024gpt4ocard} to summarize issues into underspecified variants, removing important details while preserving repository terminology. Given the automatic nature of the construction process, we performed an additional spot-check analysis to verify dataset validity (Appendix~\ref{app:dataset_construction}); concrete examples are provided in Appendix~\ref{app:underspecification_examples}.

\subsection{Agent Design}
\paragraph{Agent Framework} We use the OpenHands \citep{wang2025openhands} agent framework for all experiments. This framework enables an LLM to iteratively write, execute, and debug code within a secure sandbox environment, using tools for navigating codebases, editing files and executing Bash/Python scripts. To ensure comparability with \citet{vijayvargiya2026interactive}, all agent setups are provided with a maximum of 100 iterations.

\begin{figure}[t]
    \centering
	\includegraphics[width=\columnwidth]{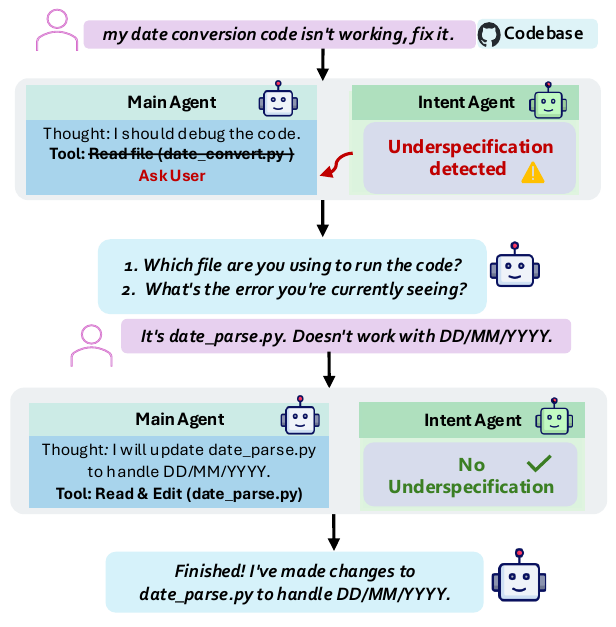}
    \caption{Illustration of the uncertainty-aware multi-agent scaffold. The Intent Agent monitors the state history at each turn, halting execution to query the user if missing information is required.}
	\label{fig:intent_agent}
\end{figure}

\paragraph{Agent Backbones} We evaluate two frontier backbone LLMs: Claude Sonnet 4.5 \citep{anthropic2025sonnet45}, a proprietary model, and Kimi K2.6 \citep{moonshot2026kimik26}, an open-weight model. Both perform competitively in many coding benchmarks, serving as strong proxies for current agentic capabilities in interactive software engineering settings.

\paragraph{User Simulator} Following recent interactive environments that utilize ``oracle'' user simulators with access to complete information (e.g., \citealp{yao2025taubench, zhou2025sweet}), we employ GPT-5.1 \citep{openai2025gpt51} as the simulated user for all interactive agent configurations. We select this model to maintain consistency with the interactive SWE-bench setting introduced by \citet{vijayvargiya2026interactive}. The simulated user is provided with the original, fully specified issue and is constrained to answer queries from the coding agent using only this withheld context. The user simulator prompt and all additional agent and task prompts are detailed in Appendices~\ref{app:task_prompts}--\ref{app:user_sim_prompt}.

\subsection{Task Design}
\subsubsection{Baselines}
Following \citet{vijayvargiya2026interactive}, we evaluate our approach on the SWE-bench Verified dataset against three distinct baseline configurations.

\paragraph{\textsc{Full}} This is the standard SWE-bench setting, where the agent is provided with a fully specified version of the GitHub issue. The agent is prohibited from interacting with a user, representing default fully autonomous agent behavior.

\paragraph{\textsc{Hidden}} In this configuration, the agent is provided with an underspecified version of the GitHub issue where details are missing. As in the \textsc{Full} baseline, the agent cannot interact with a user.

\paragraph{\textsc{Interactive Baseline}} The agent receives an underspecified version of the GitHub issue (as with the \textsc{Hidden} baseline) but can interact with a simulated user who possesses the fully specified issue. Importantly, the task prompt is modified to explicitly inform the agent that the issue description is incomplete, making it compulsory to query the user before proceeding with any execution. Consequently, this hardcoded instruction forces a predetermined conversational turn with the user rather than evaluating independent information-seeking behavior.

\subsubsection{Uncertainty-Aware Agents}
To investigate how agents can independently detect and resolve underspecification, we propose two scaffolds. We refer to these as ``uncertainty-aware'' since they are designed to adapt their information-seeking behavior based on the extent of missing context. While these agents can query the same simulated user as the \textsc{Interactive Baseline}, they do not rely on a hardcoded interaction prompt, instead using the default SWE-bench task prompt. Because they receive no prior warning that the issue is underspecified, they must independently identify missing context and query the simulated user only when they determine it is necessary.

\paragraph{\textsc{Uncertainty-Aware (Single)}}
In this configuration, a single coding agent is prompted at each turn to check for underspecification and, if detected, to query the user. We refer to the agent hereafter as \textsc{UA-Single}.

\paragraph{\textsc{Uncertainty-Aware (Multi)}}
We investigate whether leveraging a multi-agent scaffold can improve underspecification detection. By assigning specialized roles to multiple LLMs, multi-agent systems can tackle increasingly complex tasks \citep{hong2024metagpt, qian-etal-2024-chatdev, park2023generative}. As illustrated in Figure~\ref{fig:intent_agent}, to decouple code execution from underspecification detection, we design a multi-agent scaffold consisting of two agents. The \textbf{Main Agent} is equipped with standard tools to navigate repositories, edit files, and execute code. The \textbf{Intent Agent} monitors the state history at each turn to determine if the user's intent or repository context contains missing information. The Intent Agent is provided with its own dedicated system prompt (detailed in Appendix~\ref{app:scaffold_prompts}) and has access to a single tool that outputs a binary decision indicating whether clarification is required. Whenever the Intent Agent detects underspecification, the Main Agent's next action is constrained to query the user. In our experiments, both agents share the same LLM backbone. We refer to the agent hereafter as \textsc{UA-Multi}.

\begin{figure}[t]
    \centering
	\includegraphics[width=\columnwidth]{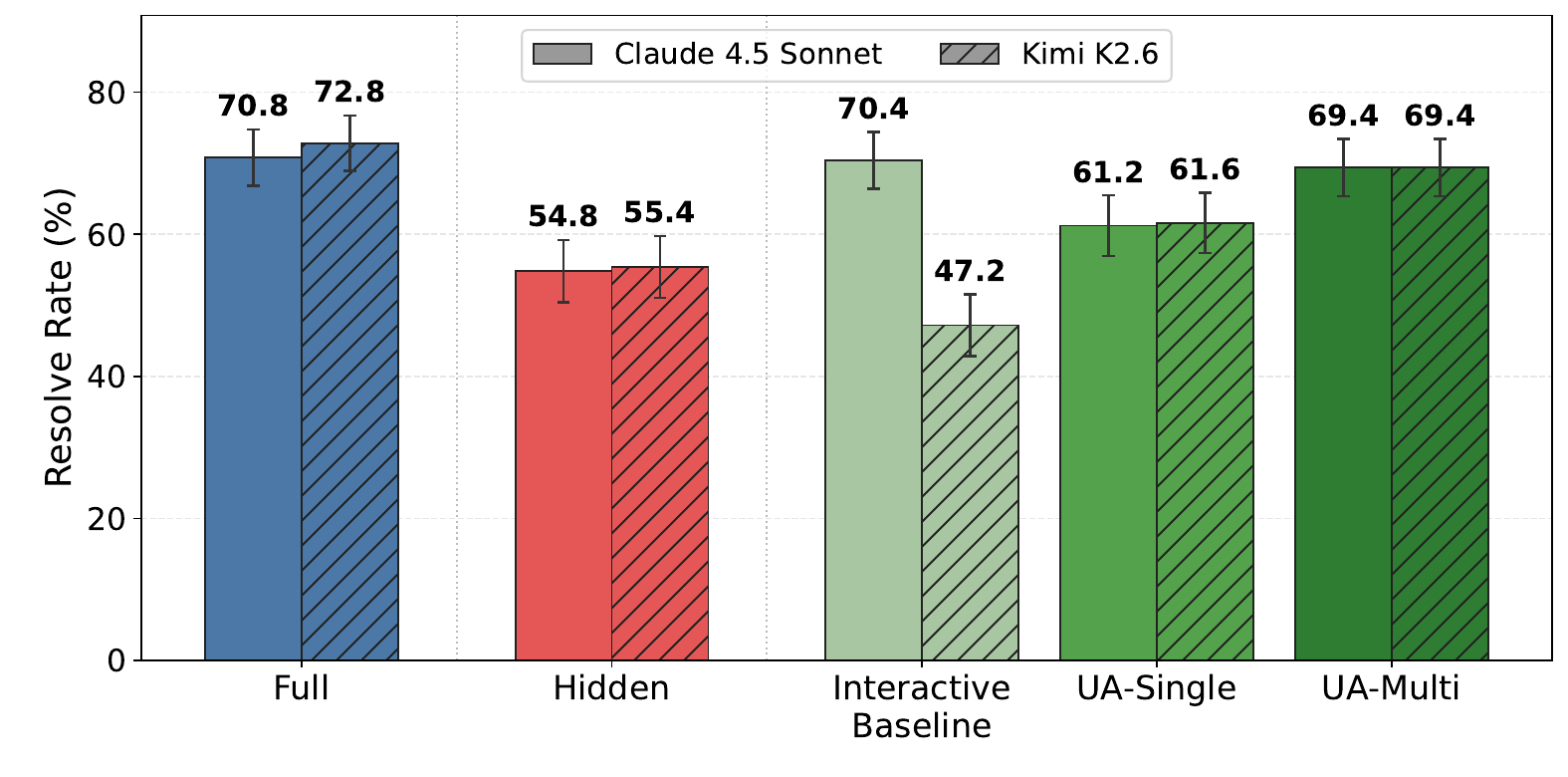}
    \caption{Task resolve rates (\%) across evaluation settings for Claude Sonnet 4.5 (solid) and Kimi K2.6 (hatched), with 95\% confidence intervals.}
	\label{fig:resolve_rates}
\end{figure}

\section{Results and Discussion}
Figure~\ref{fig:resolve_rates} presents the overall resolve rates.\footnote{Our reported resolve rates for the \textsc{Full} baselines are slightly lower than the official OpenHands results (Claude Sonnet 4.5: 70.80\% vs. 74.20\%; Kimi K2.6: 72.80\% vs. 74.60\%). This is likely due to a combination of running for a maximum of 100 iterations instead of 500, as well as minor system prompt differences. Additionally, all reported $p$-values are computed via non-parametric permutation tests.} For Claude Sonnet 4.5, \textsc{UA-Single} outperforms the \textsc{Hidden} baseline (61.20\% vs. 54.80\%) but falls short of the prompted \textsc{Interactive Baseline} (70.40\%). However, separating underspecification detection and code execution in \textsc{UA-Multi} substantially improves performance. It achieves a 69.40\% resolve rate, yielding a significant improvement over \textsc{UA-Single} ($p < 0.001$) and closing the gap with the \textsc{Interactive Baseline} ($p = 0.621$) and \textsc{Full} ($p = 0.458$) configurations.

For Kimi K2.6, we observe similar trends, with \textsc{UA-Multi} also achieving a 69.40\% resolve rate, significantly improving over \textsc{UA-Single} (61.60\%, $p < 0.001$), and approaching the performance of \textsc{Full}, although a significant gap remains ($p < 0.05$). A notable exception is the \textsc{Interactive Baseline}, which achieves the lowest resolve rate of 47.20\%. This is largely driven by instruction-following failures, where the model erroneously used the task completion tool to ask questions, highlighting the brittleness of hardcoded prompts and reliance on underlying model capability.

\paragraph{Agent Uncertainty is Calibrated to Task Difficulty} The success of \textsc{UA-Multi} can be largely attributed to its ability to discern \textit{when} to ask questions. For Claude Sonnet 4.5, while it queried the user at least once in fewer overall tasks than \textsc{UA-Single} ($N_{\text{ask}}{=}344$ vs. $N_{\text{ask}}{=}369$), its interventions were substantially more effective, resolving 65.99\% of queried tasks compared to \textsc{UA-Single}'s 55.81\% and the \textsc{Hidden} baseline's 44.48\%, while closely matching the \textsc{Interactive Baseline} and \textsc{Full} settings (both 66.57\%). Crucially, the \textsc{Interactive Baseline} achieves its high resolve rate at the cost of overclarification, asking questions in nearly every instance regardless of necessity. In contrast, \textsc{UA-Multi} is highly selective in its interaction. For the 156 tasks where \textsc{UA-Multi} refrained from asking, it still achieved a 76.92\% resolve rate, closely matching the \textsc{Hidden} baseline (77.56\%). Conversely, \textsc{UA-Multi} with Kimi K2.6 queries more than \textsc{UA-Single} ($N_{\text{ask}}{=}435$ vs. $N_{\text{ask}}{=}167$), but nevertheless achieves a competitive resolve rate on tasks where it refrains from asking (75.38\%).

Appendix~\ref{app:difficulty_ask_rates} further validates query rate calibration across human-annotated task difficulty levels for SWE-bench Verified \citep{chowdhury2024swebenchverified}, showing that \textsc{UA-Multi} demonstrates an improved ability to distinguish when to ask based on task complexity. For Claude Sonnet 4.5, \textsc{UA-Multi} exhibits a 9.28\% higher ask rate for medium (``15 min -- 1 hour'') tasks than easy (``$<$15 min'') tasks, compared to only a 2.43\% increase for \textsc{UA-Single}. For Kimi K2.6, both agents show similar easy-to-medium query rate increases, although \textsc{UA-Single}'s ask rate declines for harder tasks, suggesting its query decisions do not reliably track task difficulty.

\begin{table}[t]
\centering
\small
\begin{tabularx}{\columnwidth}{X cc}
\toprule
& \multicolumn{2}{c}{\textbf{Resolve Rate (\%)}} \\
\cmidrule(lr){2-3}
\textbf{Evaluation Setting} & \textbf{Asked} & \textbf{Did Not Ask} \\
\midrule
\multicolumn{3}{l}{\textit{Claude Sonnet 4.5} ($N_{\text{ask}}=344$, $N_{\text{not}}=156$)} \\
\quad UA-Multi & 65.99 & 76.92 \\
\quad Full & \textbf{66.57} & \textbf{80.13} \\
\quad Hidden & 44.48 & 77.56 \\
\quad UA-Single & 55.81 & 73.08 \\
\quad Interactive Baseline & \textbf{66.57} & 78.85 \\
\midrule
\multicolumn{3}{l}{\textit{Kimi K2.6} ($N_{\text{ask}}=435$, $N_{\text{not}}=65$)} \\
\quad UA-Multi & 68.51 & 75.38 \\
\quad Full & \textbf{71.95} & \textbf{78.46} \\
\quad Hidden & 52.18 & 76.92 \\
\quad UA-Single & 60.46 & 69.23 \\
\quad Interactive Baseline & 45.29 & 60.00 \\
\bottomrule
\end{tabularx}
\caption{Resolve rates (\%) conditioned on whether the \textsc{UA-Multi} agent queried the user at least once in a task. Best performing settings per subset are in \textbf{bold}.}
\label{tab:combined_ask_resolve_rate_multi}
\end{table}

\paragraph{Proactive Information Seeking} Beyond knowing \textit{when} to ask, explicitly isolating the intent-detection role improved \textit{how} the agent interacted. For Claude Sonnet 4.5, \textsc{UA-Multi} interrogated the user more iteratively, averaging 3.06 queries per task compared to 1.84 for \textsc{UA-Single}, and distributed its queries across the early and middle stages of execution. For Kimi K2.6, \textsc{UA-Multi} exhibits a much higher query rate, averaging 8.71 queries per task, with queries mainly concentrated in the middle stage of execution, suggesting poorer calibration relative to Claude Sonnet 4.5. A more detailed analysis is provided in Appendix~\ref{app:question_stats}, along with qualitative trajectory comparisons in Appendix~\ref{app:qualitative_transcripts}.

\section{Conclusion}
We investigated the abilities of LLM agents to act as collaborators by independently identifying missing information and seeking clarification in underspecified software engineering tasks. We introduced an uncertainty-aware multi-agent scaffold that isolates the role of underspecification detection. Evaluated on an underspecified variant of SWE-bench Verified, this system achieved a 69.40\% resolve rate across both proprietary and open-weight LLM backbones that effectively closed the performance gap with an autonomous agent operating on a fully specified issue.

Our results show that the multi-agent scaffold exhibits well-calibrated information-seeking behavior, accurately recognizing when an issue was already resolvable and refraining from unnecessary interaction on simpler tasks. Our results also reveal that underlying model capability shapes clarification behavior: while both LLM backbones benefit from the multi-agent scaffold, Kimi K2.6 exhibits higher query rates and poorer calibration, suggesting that robust uncertainty-aware interaction may require targeted training beyond architectural design alone. These findings present a promising step towards deploying agents not only as autonomous coding assistants, but as proactive collaborators capable of detecting and resolving external uncertainty arising from real-world underspecification.

\section*{Limitations}
\paragraph{User Simulator}
Our evaluation relies on an LLM-based user simulator to provide withheld information. While we implemented strict guardrails to prevent unintended leakage and generally observed reasonable simulator responses, recent studies highlight that LLM-simulated users can be unreliable proxies for human behavior, often being unnaturally cooperative and failing to reflect the nuance and variance of real human users \citep{naous2026flipping, seshadri2026lost}. Results may therefore vary if real human users interact with the agents that we presented in this work.

\paragraph{Prompting and Training}
While our most successful approach relies on a multi-agent scaffold with tailored prompts for each agent, it demonstrates that frontier models possess the latent capacity to monitor their own uncertainty and proactively seek clarification out of the box. Rather than relying on prompting specialized agents, future work could explore utilizing these successful interaction trajectories to train single models with standard finetuning or reinforcement learning (RL) techniques to natively exhibit this calibrated, information-seeking behavior \citep[e.g.,][]{andukuri2024stargate, bhargava2024promptbaking, sun2026training}.

\paragraph{Model Generalization and Cost} Our experiments were conducted using both a proprietary model (Claude Sonnet 4.5) and an open-weight model (Kimi K2.6), demonstrating that our multi-agent scaffold generalizes across model families. By relying on frontier models, our evaluation incurred a non-trivial financial cost (detailed in Appendix~\ref{app:computing_costs}). Furthermore, previous studies indicate that smaller open-weight models severely lack the calibration required for interactive underspecification \citep{vijayvargiya2026interactive}, even with explicit prompting. The current findings may therefore be limited to frontier models and may not directly translate to smaller open-source models.

\section*{Ethical Considerations}
Our work investigated the ability of LLM agents to detect external uncertainty and seek clarification on underspecified software engineering tasks. While our results demonstrate that uncertainty-aware scaffolds can effectively resolve underspecified GitHub issues, SWE-bench Verified represents only a subset of software engineering tasks. As such, our empirical findings should not be extrapolated to suggest that agents will reliably detect missing information in other, often high-stakes environments, such as security-critical applications without further experimentation.

We also acknowledge the environmental and financial costs associated with the deployment of multi-agent systems. As detailed in our cost breakdown (Appendix~\ref{app:computing_costs}), the increased inference overhead in multi-agent systems is a trade-off that must be weighed against the gains in developer productivity.

\paragraph{AI Use.} We used an AI assistant to assist with experimental code generation, and help improve the clarity and flow of the writing while revising the paper. However, all original writing, research conceptualization, methods, experiments, and analyses were performed by the authors, and any AI outputs were carefully verified.

\section*{Acknowledgments}
This work was supported by funding from the Vienna Science and Technology Fund (WWTF) through the project ``Understanding Language in Context'' (WWTF Vienna Research Group VRG23-007).

\bibliography{custom}

\clearpage

\appendix

\section{Dataset Construction and Validation}
\label{app:dataset_construction}

As mentioned in Section~\ref{sec:dataset}, we conduct our evaluation using an underspecified variant of SWE-bench Verified, directly adopting the dataset introduced by \citet{vijayvargiya2026interactive}. The authors automatically constructed this dataset through prompting GPT-4o to summarize the original GitHub issue. The prompt instructed the model to retain repository-specific terminology while removing important technical details.

To validate the synthetically generated dataset, \citet{vijayvargiya2026interactive} applied distributional difference analysis \citep{zhong2023goal}, an LLM-driven approach for discovering and describing differences between corpora, to compare the generated issues against naturally occurring underspecified issues in the original SWE-bench dataset. Their analysis found that natural issues typically retain more concrete technical details, such as code snippets, specific error messages, and file/line references, compared to the generated issues. Beyond this difference, however, the synthetic and natural issues were not found to be otherwise distinguishable. Consequently, the synthetic generation process produces instances that are often strictly \textit{more} underspecified than natural occurrences. For our evaluation, this characteristic of the synthetic dataset is advantageous, making the task more challenging and heavily penalizing blind execution while encouraging proactive interaction.

\paragraph{Empirical Spot-Check Analysis}
To further verify the dataset's validity for our evaluation, we conducted a manual qualitative analysis on a random sample of 10 task instances. For each instance, we compared the original, fully specified issue description with the synthetically generated underspecified issue description to determine whether the removed information was necessary for solving the task.

In 5 instances, the removed information was essential for solving the task, such as specific OS configurations or user-provided value constraints. In the remaining 5 instances, the summarized description retained sufficient technical details (e.g., specific class or method names) to likely allow an agent to successfully localize the relevant files in the repository requiring edits to correctly resolve the issue. Figure~\ref{fig:specification_examples} illustrates one example of each case.

Notably, this variance closely mirrored actual agent performance. For instance, the \textsc{Hidden} baseline using Claude Sonnet 4.5 failed to solve all 5 instances where our manual analysis revealed that essential information was removed, but successfully solved the 5 instances where sufficient technical details were retained. This pattern is consistent with the significant performance gap between the 
\textsc{Hidden} and \textsc{Full} baselines (54.80\% vs. 70.80\% for the Claude Sonnet 4.5 agent), highlighting that the dataset provides meaningful signal for when interaction is required to resolve underspecification.

\section{Task and Scaffold Prompts}
\subsection{Task Prompts}
\label{app:task_prompts}
Figure~\ref{fig:prompt_swe_combined} presents the SWE-bench task prompt template. The \textsc{Full} baseline is provided with this prompt including the original, fully specified issue. The \textsc{Hidden} baseline and both \textsc{Uncertainty-Aware} scaffolds receive it with the underspecified issue. Conversely, only the \textsc{Interactive Baseline} receives the augmented variant (highlighted in bold in Figure~\ref{fig:prompt_swe_combined}), which explicitly mentions that there is missing information and instructs the agent to first ask questions before proceeding.

\begin{figure*}[ht]
\centering
\begin{tcolorbox}[
  enhanced,
  sharp corners,
  colback=white,
  colframe=black,
  title=SWE-bench Task Prompt,
  fonttitle=\bfseries,
  boxrule=0.5pt,
  width=\textwidth
]
\footnotesize
\input{figure/swe_agent_interact_prompt}
\end{tcolorbox}
\caption{The base SWE-bench task prompt provided to the agents. The highlighted block in bold indicates the explicit clarification instructions added in the interactive variant, provided \textit{only} to the \textsc{Interactive Baseline}.}
\label{fig:prompt_swe_combined}
\end{figure*}

\subsection{Agent Scaffold Prompts}
\label{app:scaffold_prompts}
Figure~\ref{fig:agent_scaffolds} shows specific prompts required for the uncertainty-aware agents.

For \textsc{Uncertainty-Aware (Single)}, we provide a recurring reminder prompt to the agent at each turn (Figure~\ref{fig:agent_scaffolds}A) to assess for underspecification. If the agent detects underspecification, it is encouraged to use the \texttt{clarify} tool to query the user.

For \textsc{Uncertainty-Aware (Multi)}, we employ a customized system prompt for the Intent Agent (Figure~\ref{fig:agent_scaffolds}B). This prompt instructs the agent to analyze the state history at each turn to detect underspecification. Rather than performing code edits, the Intent Agent is constrained to identifying missing information and determining when the Main Agent must pause execution to seek clarification from the simulated user.

To reduce the over-querying behavior observed with Kimi K2.6, we additionally experimented with an augmented Intent Agent prompt that instructs the agent not to flag underspecification when the user has already indicated they cannot provide further details. Specifically, we appended the following condition to the Intent Agent prompt ``... or when the user cannot provide any further details or explicitly asks you to keep working on the task''. While this modification reduced the number of times the agent asked a user while solving individual tasks, the agent continued to flag the vast majority of tasks as underspecified. This over-querying behavior, i.e., the tendency to ask questions even when it would be possible to solve the task without further clarification, suggests that the underlying Kimi K2.6 model has a limited ability to follow instructions and determine when tasks are underspecified.

\begin{figure*}[ht]
\centering
\begin{tcolorbox}[
  enhanced,
  sharp corners,
  colback=white,
  colframe=black,
  title=Uncertainty-Aware Agent Scaffold Prompts,
  fonttitle=\bfseries,
  boxrule=0.5pt,
  width=\textwidth
]
\footnotesize

\textbf{A. \textsc{Uncertainty-Aware (Single)} Reminder Prompt}
\vspace{0.1cm}
\hrule
\vspace{0.1cm}
\input{figure/single_agent_reminder_prompt}

\vspace{0.4cm}
\textbf{B. \textsc{Uncertainty-Aware (Multi)} Intent Agent Prompt}
\vspace{0.1cm}
\hrule
\vspace{0.1cm}
\input{figure/intent_agent_prompt}

\end{tcolorbox}
\caption{Prompts required for the custom uncertainty-aware agent scaffolds. Part A shows the reminder prompt for the \textsc{Uncertainty-Aware (Single)} agent at each turn. Part B shows the system prompt for the specialized Intent Agent in the \textsc{Uncertainty-Aware (Multi)} agent scaffold.}
\label{fig:agent_scaffolds}
\end{figure*}

\subsection{User Simulator Prompt}
\label{app:user_sim_prompt}
Figure~\ref{fig:user_simulator_prompt} shows the prompt provided to the user simulator, adapted from the prompt in \citet{vijayvargiya2026interactive}. Initial experiments revealed that the simulated user occasionally provided misaligned guidance 
due to a lack of awareness regarding the specific constraints of our evaluation environment (i.e., OpenHands agent in SWE-bench). To prevent the simulator from misleading the agent or leaking unintended information, we augmented the original prompt with the following environment-specific guardrails:

\begin{figure*}[ht]
\centering
\begin{tcolorbox}[
  enhanced,
  sharp corners,
  colback=white,
  colframe=black,
  title=User Simulator Prompt,
  fonttitle=\bfseries,
  boxrule=0.5pt,
  width=\textwidth
]
\footnotesize
\vspace{0.1cm}
\input{figure/user_simulator_prompt}
\end{tcolorbox}
\caption{The prompt provided to the user simulator, including specific guardrails to prevent unintended test modifications (rule 5) and resolve environment directory mismatches (rule 6).}
\label{fig:user_simulator_prompt}
\end{figure*}

\begin{itemize}
    \item \textbf{Preventing test modifications (Rule 5):} In SWE-bench, test files should not be modified by the agent. Because the original simulated user lacked this context, it failed to correct the agent when it attempted to edit files inside the \texttt{/testbed} directory. Rule 5 was added to instruct the simulated user to explicitly remind the agent of this if it asked about changing test files.
    \item \textbf{Resolving directory mismatches (Rule 6):} Agents in OpenHands operate within a designated \texttt{/workspace} directory, whereas SWE-bench imports and runs tests in a separate \texttt{/testbed} directory. Agents often became confused when changes in \texttt{/workspace} were not reflected in the tests run in \texttt{/testbed}. Rule 6 was added to instruct the simulated user to explicitly remind the agent of this if it asked why its edits weren't being reflected.
\end{itemize}

Additionally, to prevent unbounded interactions and encourage efficient information-seeking, the simulated user is restricted to three interaction turns per trajectory. After this limit is reached, it issues a canned response instructing the agent to proceed with the available information. Note that the \textsc{Interactive Baseline} configuration is explicitly warned of this limit upfront, while our uncertainty-aware agents are not, testing their ability to independently seek information in an efficient manner.

\section{Question Analyses}
\subsection{Question Statistics}
\begin{table*}[t]
\centering
\small
\setlength{\tabcolsep}{5pt}
\begin{tabular}{l ccc cc ccc}
\toprule
& \multicolumn{3}{c}{\textbf{Interaction Volume}} & \multicolumn{2}{c}{\textbf{Token Length}} & \multicolumn{3}{c}{\textbf{Timing Distribution (\%)}} \\
\cmidrule(lr){2-4} \cmidrule(lr){5-6} \cmidrule(lr){7-9}
\textbf{Agent} & \makecell{\textbf{\# Queried} \\ \textbf{Tasks}} & \makecell{\textbf{Total} \\ \textbf{Queries}} & \makecell{\textbf{Avg \#} \\ \textbf{Q/Task}} & \textbf{Avg Q} & \textbf{Avg A} & \textbf{Early} & \textbf{Mid} & \textbf{Late} \\
\midrule
\multicolumn{9}{l}{\textit{Claude Sonnet 4.5}} \\
\quad UA-Multi & 344 & 1053 & 3.06 & 171.57 & 173.65 & 41.8 & 43.4 & 14.8 \\
\quad UA-Single & 369 & 679 & 1.84 & 181.35 & 229.76 & 25.0 & 31.1 & 43.9 \\
\quad Interactive Baseline & 496 & 508 & 1.02 & 251.33 & 415.53 & 97.6 & 1.4 & 1.0 \\
\midrule
\multicolumn{9}{l}{\textit{Kimi K2.6}} \\
\quad UA-Multi & 435 & 3791 & 8.71 & 96.73 & 57.06 & 20.7 & 46.5 & 32.8 \\
\quad UA-Single & 167 & 206 & 1.23 & 143.18 & 247.97 & 51.0 & 32.5 & 16.5 \\
\quad Interactive Baseline & 194 & 194 & 1.00 & 176.63 & 244.07 & 28.4 & 66.5 & 5.2 \\
\bottomrule
\end{tabular}
\caption{Question statistics across evaluation settings. \textit{\# Queried Tasks} denotes the number of tasks where the agent initiated at least one query. Timing distributions are categorized into Early (1st--3rd decile), Mid (4th--7th decile), and Late (8th--10th decile) based on the question's event ID position within the trajectory.}
\label{tab:combined_question_stats}
\end{table*}
\label{app:question_stats}
Table~\ref{tab:combined_question_stats} provides a detailed breakdown of clarification-seeking behavior across the three interactive agent configurations using both LLM backbones.

Regarding the amount of interaction, the \textsc{Interactive Baseline} initiates queries in nearly all tasks for Claude Sonnet 4.5 ($N_\text{ask}{=}496$, 1.02 Q/Task), but far fewer for Kimi K2.6 ($N_\text{ask}{=}194$, 1.00 Q/Task). This reduction is largely driven by instruction-following failures, where the model erroneously used the task completion tool to ask questions rather than initiating a proper query turn, resulting in many tasks where interaction was attempted but not successfully executed.

In general, the uncertainty-aware agents demonstrate more conservative query initiation but greater query volume. \textsc{UA-Multi} initiated queries in 344 tasks for Claude Sonnet 4.5 and 435 tasks for Kimi K2.6, engaging in significantly more iterative dialogue (Claude: 3.06 Q/Task; Kimi: 8.71 Q/Task). The much higher query rate for Kimi K2.6 also reflects the model's tendency to over-flag underspecification. \textsc{UA-Single} initiated queries in more tasks with Claude Sonnet 4.5 ($N_\text{ask}{=}369$) but in fewer tasks with Kimi K2.6 ($N_\text{ask}{=}167$). For both models, the query rate was substantially lower than for the multi-agent scaffold (Claude: 1.84 Q/Task; Kimi: 1.23 Q/Task).

The length of these interactions also varies. For Claude Sonnet 4.5, the \textsc{Interactive Baseline} generates highly verbose queries (251.33 tokens) and receives correspondingly long answers (415.53 tokens), reflecting a strategy to ask for all potentially missing information upfront.
In contrast, \textsc{UA-Multi} uses the most concise queries (171.57 tokens) and receives the shortest answers (173.65 tokens), supporting the observation that the multi-agent scaffold asks highly targeted, context-specific questions based on intermediate tool observations (e.g., see Figure~\ref{fig:qualitative_transcripts}). \textsc{UA-Single} falls between the two (181.35/229.76 tokens). For Kimi K2.6, the pattern differs: \textsc{UA-Multi} produces shorter queries (96.73 tokens) but receives markedly shorter answers (57.06 tokens), suggesting lower-quality clarification requests that elicit less informative responses. Notably, \textsc{UA-Single} with Kimi K2.6 asks fewer but more substantive questions (143.18 tokens) and receives considerably longer answers (247.97 tokens).

Finally, we analyze the temporal distribution of these queries. We categorize the timing of each question based on its event ID position within the agent trajectory, divided into deciles: ``Early'' (1st--3rd decile), ``Mid'' (4th--7th decile), and ``Late'' (8th--10th decile). For Claude Sonnet 4.5, the \textsc{Interactive Baseline} concentrates almost all questions in the Early stage (97.6\%), while \textsc{UA-Single} skews heavily Late (43.9\%), suggesting it often attempts code modifications before recognizing an information gap. In contrast, \textsc{UA-Multi} distributes queries across Early (41.8\%) and Mid (43.4\%) stages, reflecting continuous monitoring throughout the trajectory. For Kimi K2.6, the temporal patterns differ: \textsc{UA-Multi} concentrates queries in the Mid stage (46.5\%), while \textsc{UA-Single} skews Early (51.0\%). Notably, the \textsc{Interactive Baseline} for Kimi K2.6 is heavily concentrated in the middle of the trajectory (66.5\%), in contrast to the Early concentration seen with Claude Sonnet 4.5, likely reflecting the instruction-following failures that disrupted early-stage querying.

\subsection{Query Frequency by Task Difficulty}

\begin{table}[t]
\centering
\small
\begin{tabular}{lrrr}
\toprule
& & \multicolumn{2}{c}{\textbf{Ask Rate (\%)}} \\
\cmidrule(lr){3-4}
\textbf{Difficulty Level} & \textbf{\# Tasks} & \textbf{\textsc{UA-Single}} & \textbf{\textsc{UA-Multi}} \\
\midrule
\multicolumn{4}{l}{\textit{Claude Sonnet 4.5}} \\
$<$15 min fix & 194 & 71.13 & 62.37 \\
15 min--1 hour & 261 & 73.56 & 71.65 \\
1--4 hours & 42 & 85.71 & 78.57 \\
$>$4 hours & 3 & 100.00 & 100.00 \\
\midrule
\multicolumn{4}{l}{\textit{Kimi K2.6}} \\
$<$15 min fix & 194 & 29.90 & 82.99 \\
15 min--1 hour & 261 & 36.02 & 88.89 \\
1--4 hours & 42 & 30.95 & 92.86 \\
$>$4 hours & 3 & 66.67 & 100.00 \\
\bottomrule
\end{tabular}
\caption{Ask rates by SWE-bench Verified difficulty (estimated time-to-fix) for uncertainty-aware (UA) agents.}
\label{tab:combined_difficulty_ask_rates}
\end{table}
\label{app:difficulty_ask_rates}
To evaluate whether agents appropriately calibrate their uncertainty to the complexity of the task, we mapped their querying behavior to the human-annotated difficulty levels provided by SWE-bench Verified \citep{chowdhury2024swebenchverified}. Difficulty is categorized by the estimated time required for a human developer to fix the issue.

Table~\ref{tab:combined_difficulty_ask_rates} presents the ask rates for both \textsc{UA-Single} and \textsc{UA-Multi}. For Claude Sonnet 4.5, \textsc{UA-Multi} exhibits a greater difference in ask rates than \textsc{UA-Single}, particularly between the ``$<$15 min fix'' and ``15 min -- 1 hour'' levels (9.28\% vs. 2.43\% increase). Notably, both scaffolds exhibit an increased ask rate for the harder tasks with both agents reaching a 100\% interaction rate on the most difficult ``$>$4 hours'' tasks. For Kimi K2.6, \textsc{UA-Multi} queries the vast majority of tasks at every difficulty level (82.99\%--100\%) with a modest monotonic increase, although the consistently high query rate suggests reduced sensitivity to task complexity. \textsc{UA-Single} displays a non-monotonic pattern, with ask rates declining from medium (36.02\%) to harder tasks (30.95\%), suggesting its query decisions do not reliably track task difficulty.

\subsection{Conditional Resolve Rates for \textsc{Uncertainty-Aware (Single)}}
\begin{table}[t]
\centering
\small
\begin{tabularx}{\columnwidth}{X cc}
\toprule
& \multicolumn{2}{c}{\textbf{Resolve Rate (\%)}} \\
\cmidrule(lr){2-3}
\textbf{Evaluation Setting} & \textbf{Asked} & \textbf{Did Not Ask} \\
\midrule
\multicolumn{3}{l}{\textit{Claude Sonnet 4.5} ($N_{\text{ask}}=369$, $N_{\text{not}}=131$)} \\
\quad UA-Single & 58.54 & 68.70 \\
\quad Full & 67.48 & \textbf{80.15} \\
\quad Hidden & 47.43 & 75.57 \\
\quad UA-Multi & 66.40 & 77.86 \\
\quad Interactive Baseline & \textbf{68.29} & 76.34 \\
\midrule
\multicolumn{3}{l}{\textit{Kimi K2.6} ($N_{\text{ask}}=167$, $N_{\text{not}}=333$)} \\
\quad UA-Single & 60.48 & 62.16 \\
\quad Full & \textbf{69.46} & \textbf{74.47} \\
\quad Hidden & 35.33 & 65.47 \\
\quad UA-Multi & 65.87 & 71.17 \\
\quad Interactive Baseline & 37.72 & 51.95 \\
\bottomrule
\end{tabularx}
\caption{Resolve rates (\%) conditioned on whether the \textsc{UA-Single} agent queried the user at least once. $N_{\text{ask}}$ and $N_{\text{not}}$ denote subset sizes. Best per-subset results in \textbf{bold}.}
\label{tab:combined_ask_resolve_rate_single}
\end{table}

Table~\ref{tab:combined_ask_resolve_rate_single} details the task resolve rates conditioned on whether the \textsc{UA-Single} agent chose to interact with the user. For Claude Sonnet 4.5, when \textsc{UA-Single} chose to ask questions (369 of 500 tasks), it successfully elicited missing information, resolving 216 tasks compared to only 175 for the \textsc{Hidden} baseline on the same subset. Conversely, on the 131 tasks where the agent refrained from interaction, its performance closely mirrored the \textsc{Hidden} baseline (90 vs. 99 resolved tasks). These results indicate that \textsc{UA-Single} with Claude Sonnet 4.5 can effectively detect when missing information is critical for task resolution, although the performance gap with other configurations suggests that its clarification-seeking behavior remains less calibrated compared to \textsc{UA-Multi}. For Kimi K2.6, the pattern is weaker: resolve rates are nearly identical regardless of whether the agent asked (60.48\%) or did not (62.16\%), suggesting more weakly calibrated clarification-seeking behavior.

\subsection{Qualitative Example of Agent Interaction}
\begin{figure*}[ht]
\centering

\begin{tcolorbox}[colback=gray!5, colframe=gray!50, rounded corners, width=\textwidth, title=\textsc{Interactive Baseline}, coltitle=black, fonttitle=\bfseries, top=2pt, bottom=2pt]
\footnotesize
\textbf{Agent:} Based on the issue description mentioning ``Python expression'', ``assertion failure in the Python compiler'', and ``constant names'', this is likely related to pytest's assertion rewriting mechanism. Let me ask specific questions:
\begin{enumerate}
    \item What is the \textbf{specific Python expression} that causes the crash? 
    \item What is the \textbf{exact error message} or assertion failure from the Python compiler? Is there a stack trace or specific assertion message that appears?
    \item Is there a \textbf{specific Python bug report} or bpo issue number that this is related to?  The description mentions ``a known issue'' - what is the reference for this?
\end{enumerate}
\hdashrule{\textwidth}{0.5pt}{1pt}
\vspace{0.1cm}
\textbf{User:}
\vspace{-0.2cm}
\begin{enumerate}
    \item The minimal reproducer is this call into pytest's mark expression compiler: \texttt{Expression.compile("False")}. Just compiling the string "False" (as a mark expression) is enough to trigger the crash on a debug build of Python 3.8+.
    \item The crash is an assertion failure in the CPython compiler. On a debug build, it looks like:
\begin{lstlisting}[
  basicstyle=\ttfamily\footnotesize,
  breaklines=true,
  showstringspaces=false,
  columns=fullflexible
]
python: Python/compile.c:3559: compiler_nameop: Assertion
    `!_PyUnicode_EqualToASCIIString(name, "None") &&
     !_PyUnicode_EqualToASCIIString(name, "True") &&
     !_PyUnicode_EqualToASCIIString(name, "False")' failed.
[1]    29440 abort (core dumped)  python
\end{lstlisting}
    \item The related CPython bug report is bpo-40870: \url{https://bugs.python.org/issue40870}. That's the known issue aimed at improving this behavior I was referring to.
\end{enumerate}
\end{tcolorbox}

\vspace{0.1cm} 

\begin{tcolorbox}[colback=gray!5, colframe=gray!50, rounded corners, width=\textwidth, title=\textsc{Uncertainty-Aware (Single)}, coltitle=black, fonttitle=\bfseries, top=2pt, bottom=2pt]
\footnotesize
\textbf{Agent:} I need some clarification before proceeding:
\begin{enumerate}
    \item What is the specific Python expression that causes the interpreter to crash on Python 3.8+ debug builds?  For example, is it \textbf{related to assertions, specific constant values, or particular syntax patterns}?
    \item Which constant names are problematic and cause the assertion failure in the Python compiler? For example, are they names like, \textbf{\texttt{'\_\_debug\_\_'}, \texttt{'True'}, \texttt{'False'}, \texttt{'None'}}, or other special constants? 
    \item You mentioned you've already taken care of test file changes. Are these \textbf{test files already committed in the repository}, or should I look for them in a specific location? Can you point me to the test file that demonstrates this issue?
    \item Is there a \textbf{specific GitHub issue number} or Python bug tracker reference (bpo number) for the 'known issue' you mentioned that is related to this problem?
\end{enumerate}
\hdashrule{\textwidth}{0.5pt}{1pt}
\vspace{0.1cm}
\textbf{User:}
\vspace{-0.2cm}
\begin{enumerate}
    \item The minimal expression is just using \texttt{Expression.compile("False")}  (the string "False" as the expression). That alone triggers the crash on a Python 3.8+ debug build.
    \item The problematic constant names are \texttt{None}, \texttt{True}, and \texttt{False}. The assertion in \texttt{compiler\_nameop} explicitly checks that the name is not any of those three.
    \item The tests for this issue are already handled; you shouldn't modify any test files further.
    \item Yes, the related Python bug tracker reference is bpo-40870 (\url{https://bugs.python.org/issue40870}).
\end{enumerate}
\end{tcolorbox}

\caption{Comparison of user-agent interactions on the \texttt{pytest-dev\_\_pytest-7324} task (Claude Sonnet 4.5). Both the \textsc{Interactive Baseline} and the \textsc{Uncertainty-Aware (Single)} agent successfully identify missing information but group all their queries into a single, upfront interaction turn. Bold text highlights \textsc{Interactive Baseline}'s generic queries versus the more specific, technical queries of \textsc{Uncertainty-Aware (Single)}.}
\label{fig:qualitative_transcripts}
\end{figure*}

\begin{figure*}[ht]
\ContinuedFloat
\centering

\begin{tcolorbox}[colback=gray!5, colframe=gray!50, rounded corners, width=\textwidth, title=\textsc{Uncertainty-Aware (Multi)}, coltitle=black, fonttitle=\bfseries, top=2pt, bottom=2pt]
\footnotesize
\textbf{Agent (Turn 1 - Early Stage):} I need more specific information to understand and fix the issue properly:
\begin{enumerate}
    \item Can you provide the specific Python expression that causes the crash? For example, is it something like \textbf{\texttt{assert \_\_debug\_\_} or a specific constant name}?
    \item Do you have the \textbf{actual error message} or assertion failure from Python's compiler/interpreter that appears on a debug build?
    \item Is there a \textbf{specific test file or test case} that was added to reproduce this issue?
    \item What is the \textbf{'known issue'} this is linked to? Is it a Python bug tracker issue or a GitHub issue?
\end{enumerate}
\hdashrule{\textwidth}{0.5pt}{1pt}
\vspace{0.1cm}
\textbf{User:}
\vspace{-0.2cm}
\begin{enumerate}
    \item The specific expression is just \texttt{Expression.compile("False")}, i.e. the string "False" being compiled by \texttt{Expression.compile} in \texttt{src/\_pytest/mark/expression.py}.
    \item Yes, on a debug build it crashes with:
\begin{lstlisting}[
  basicstyle=\ttfamily\footnotesize,
  breaklines=true,
  showstringspaces=false,
  columns=fullflexible
]
python: Python/compile.c:3559: compiler_nameop: Assertion
    `!_PyUnicode_EqualToASCIIString(name, "None") &&
     !_PyUnicode_EqualToASCIIString(name, "True") &&
     !_PyUnicode_EqualToASCIIString(name, "False")' failed.
[1]    29440 abort (core dumped)  python
\end{lstlisting}
    \item There is a test covering this scenario, but all changes to the test files are already handled and shouldn't be modified further for this task.
    \item The linked known issue is a Python bug tracker issue: bpo-40870 (\url{https://bugs.python.org/issue40870}).
\end{enumerate}

\vspace{0.2cm}
\hrule
\vspace{0.2cm}

\textbf{Agent (Turn 2 - Mid Stage, Execution Phase):} The \textbf{test still fails after modifying the source code}. This suggests either a caching issue or the module hasn't been reloaded. Let me check if the changes were saved correctly and try reloading.
\begin{enumerate}
    \item \textbf{Should I continue debugging} why the changes aren't taking effect?
\end{enumerate}
\hdashrule{\textwidth}{0.5pt}{1pt}
\vspace{0.1cm}
\textbf{User:} No, you don't need to keep debugging that part. Remember that the tests are imported and run from \texttt{/testbed}, but your final code changes should be made in \texttt{/workspace}. If your fixes aren't reflected in the tests, it's likely due to that setup rather than a caching issue.
\end{tcolorbox}
\caption{(continued). Unlike the single-agent setups, the multi-agent framework successfully engages in additional late-stage, iterative clarification following a test failure on the \texttt{pytest-dev\_\_pytest-7324} task.}
\end{figure*}
\label{app:qualitative_transcripts}

The transcripts in Figure~\ref{fig:qualitative_transcripts} illustrate the distinct information-seeking strategies employed by the agents using Claude Sonnet 4.5 for the \texttt{pytest-dev\_\_pytest-7324} task, which involves resolving a \texttt{pytest}-related Python interpreter crash. As highlighted in bold, while the baseline relies on more generic questions, both uncertainty-aware agents ask more specific, technical questions. In particular, note how the multi-agent scaffold uniquely engages in a mid-trajectory follow-up query after observing a test failure.

\section{Underspecified Issue Examples}
\label{app:underspecification_examples}

Figure~\ref{fig:specification_examples} provides a comparison of the original and underspecified issue descriptions for two tasks. In creating the underspecified examples, the summarization process generally targets the removal of specific details such as code snippets, file paths/line references and stack traces, while preserving relevant terminology to describe the issue at a high level. However, as noted in our spot-check analysis (Appendix~\ref{app:dataset_construction}), the extent of this reduction varies across instances. In some cases, the retained high-level description still provides sufficient signal for an agent to localize the relevant code, while in others the removed details are essential for correct resolution, providing meaningful signal for when clarification-seeking behavior is required.

\begin{figure*}[ht]
\centering
\begin{tcolorbox}[
  enhanced, sharp corners, colback=white, colframe=black,
  title=\texttt{pylint-dev\_\_pylint-7080}, fonttitle=\bfseries\small,
  boxrule=0.5pt, width=\textwidth
]
\footnotesize
\input{figure/example_pylint}
\end{tcolorbox}

\vspace{0.3cm}

\begin{tcolorbox}[
  enhanced, sharp corners, colback=white, colframe=black,
  title=\texttt{scikit-learn\_\_scikit-learn-26323}, fonttitle=\bfseries\small,
  boxrule=0.5pt, width=\textwidth
]
\footnotesize
\input{figure/example_scikit-learn}
\end{tcolorbox}
\caption{Comparison of original (fully specified) and synthetically underspecified descriptions for two evaluation instances. In \texttt{pylint-dev\_\_pylint-7080} (top), the missing OS information renders completion without clarification challenging. In \texttt{scikit-learn\_\_scikit-learn-26323} (bottom), preserved class and method names are sufficient for an agent to localize and resolve the issue autonomously.}
\label{fig:specification_examples}
\end{figure*}

\section{Computing Costs}
\begin{table*}[t]
\centering
\small
\begin{tabular}{l rr rr}
\toprule
& \multicolumn{2}{c}{\textbf{Claude Sonnet 4.5}} & \multicolumn{2}{c}{\textbf{Kimi K2.6}} \\
\cmidrule(lr){2-3} \cmidrule(lr){4-5}
\textbf{Setting} & \makecell{\textbf{Total} \\ \textbf{Cost (\$)}} & \makecell{\textbf{Avg Cost} \\ \textbf{/ Task (\$)}} & \makecell{\textbf{Total} \\ \textbf{Cost (\$)}} & \makecell{\textbf{Avg Cost} \\ \textbf{/ Task (\$)}} \\
\midrule
Full & 817.02 & 1.63 & 316.18 & 0.63 \\
Hidden & 899.43 & 1.80 & 418.26 & 0.84 \\
Interactive Baseline & 697.88 & 1.40 & 270.42 & 0.54 \\
UA-Single & 1017.34 & 2.03 & 323.27 & 0.65 \\
UA-Multi & 1748.08 & 3.50 & 498.36 & 1.00 \\
\bottomrule
\end{tabular}
\caption{API costs (in USD) across evaluation settings on SWE-bench Verified.}
\label{tab:combined_cost}
\end{table*}

\label{app:computing_costs}
Table~\ref{tab:combined_cost} details the total inference costs for each evaluation setting across both LLM backbones. For Claude Sonnet 4.5, all agent inference calls were executed via the Anthropic API. While the multi-agent scaffold more than doubles the total inference cost compared to the single-agent baselines, the absolute financial cost per task remains negligible. We argue that this increase in compute expenditure is a highly favorable trade-off; by proactively resolving underspecification and significantly increasing the overall task resolve rate, the multi-agent setup ultimately saves substantial human developer time and effort that would otherwise be spent on debugging solutions that are misaligned with the original intent. For Kimi K2.6, all agent inference calls were executed via the Moonshot API. Kimi K2.6 runs at roughly 30--50\% of the cost of Claude Sonnet 4.5, offering substantial cost efficiency gains while maintaining competitive performance. Additionally, the multi-agent scaffold has a total inference cost less than double that of the single-agent baselines, making Kimi K2.6 a cost-efficient alternative for deploying the framework.

\end{document}